\documentclass{article}
\usepackage{spconf,amsmath}
\usepackage{makecell}
\usepackage{CJKutf8}
\usepackage{booktabs}
\usepackage{amsfonts}
\usepackage{graphicx}
\usepackage[colorlinks,
            linkcolor=blue,
            anchorcolor=blue,
            citecolor=blue]{hyperref}
\usepackage[marginal]{footmisc}

\newcommand{\wan}[1]{\textcolor{black}{#1}}

\title{Self-consistent Reasoning For Solving Math Word Problems}
\name{Jing Xiong$^{1}$, Zhongwei Wan${^{2,3}}$, Xiping Hu$^{1}$, Min Yang${^{2,3}}$, Chengming Li$^{1}$}
\address{
    $^1$ Sun Yat-Sen University, China \\
   $^2$ University of Chinese Academy of Sciences, China\\
   $^3$ SIAT, Chinese Academy of Sciences, China\\
    \small{
        xiongj69@mail2.sysu.edu.cn, 
        \{huxiping, lichengming\}@mail.sysu.edu.cn,
        \{zw.wan1, min.yang\}@siat.ac.cn
    }
}

\begin{document}
\maketitle
\ninept
\begin{abstract}

Math word problems (MWPs) is a task that automatically derives solution expression from a giving math problems in text. The previous studies suffer from spurious correlations between input text and output expression. To mitigate this issue, we propose a self-consistent reasoning framework called \textbf{SCR}, which attempts to adopt a pruning strategy to correct the output distribution shift so as to implicitly fix those spurious correlative samples. Specifically, we firstly obtain a sub-network by pruning a roberta2tree model, for the sake to use the gap on output distribution between the original roberta2tree model and the pruned sub-network to expose spurious correlative samples. Then, we calibrate the output distribution shift by applying symmetric Kullback-Leibler divergence to alleviate spurious correlations. In addition, SCR generates equivalent expressions, thereby, capturing the original text's logic rather than relying on hints from original text. Extensive experiments on two large-scale benchmarks demonstrate that our model substantially outperforms the strong baseline methods.

\end{abstract}

\begin{keywords}
Math Word Problems, 
Spurious correlative samples,
Pruning,
Self-consistency

\end{keywords}
\section{Introduction}
Math word problems (MWPs) \cite{bobrow1964natural} is a challenging symbolic logical reasoning task based on natural language description and draws much attention from researchers about the reasoning power of large language models \cite{wei2022chain, wang2022self, li2022advance, Wan2022GMAPGM, wang2024iot, wan2023efficient} recently. MWPs aims to automatically solve mathematical questions given in a natural language, which requires the model not only to understand the natural language but also to have the ability to reason logically. Table \ref{tab:case0} shows several examples of MWPs.


At present, there are mainly three paradigms of models that have achieved excellent performance, namely seq2seq \cite{shen2021generate,wan2023text, huang2021recall, wang2018translating, wan2023spatio, liu2023etp}, seq2tree \cite{zhang2020graph, xiong2022expression}, and complex relation extraction \cite{jie2022learning}. But all three paradigm model suffers from spurious correlations\cite{patel2021nlp,kumar2021adversarial,jie2022learning}. \wan{Take an example of Table 1, some of the previous works may obtain the same mathematical formula as "$a \div b \times c $" for Problem 1 and Problem 2, due to the similar semantic context information, e.g., \textit{Calculate the money situation.} However, if the models ignore the spurious correlations, they intend to generate the wrong solution expression for Problem 3, which has very analogous semantic information like the important words "money", "bank", and "account," which exist in problems 1 and 2. To be specific, the models that learn spurious information among problems 1 to 3 are more likely to generate the wrong expression "12500$\div$ 5 $\%$ $\times 15 \%$" instead of "12500$\div$ ( 5$\%$ + 15$\%$)" for problem 3 \textit{Calculate the money of the account.}}

Some recent models address this problem by using variational information bottlenecks\cite{xiong2022expression}. Our article considers this problem from the perspective of memorization. Some recent articles have revealed that pruning can make the model forget some hard-to-memorize samples\cite{hooker2019compressed}. In addition, \cite{jiang2021self} have revealed that long-tailed samples are easily forgettable. Usually, these long-tailed samples easily confuse the model, and the model will generate the final result based on some shallow hints. A natural hypothesis is that some spurious correlative samples are tougher for the model to learn well due to shortcuts, and these samples can be adaptively exposed by pruning \cite{jiang2021self}. A key question in the task of MWPs is how to implicitly emphasize their shortcuts between expressions and original texts when exposing spurious correlative samples through pruning. Some work about the reasoning ability of large models has also revealed that encouraging the model to produce self-consistent outputs can effectively improve reasoning performance when the model produces multiple inferences \cite{wei2022chain, wang2022self, li2022advance}. However, their work uses voting to encourage self-consistency, which cannot adaptively correct the shortcuts of expressions and original texts online through the loss function.

\begin{table}
\small
\centering
\begin{tabular}{|p{7cm}|}
\hline

{\textbf{Problem 1:} Tom takes the money from his bank account and has taken 240 dollars of his account for 3 days. If he takes the same amount of money every day, how much money will Tom take for next 2 days?}\\

\hline
{\textbf{Solution Expression:} 240$\div$3$\times$2 \quad \textbf{Solution: }160}\\ 
\hline

{\textbf{Problem 2:} Sherry has deposited 6000 dollars to the bank for the last 5 months. If she saves the same monthly money, how much will she add to the account in the next 3 months?}\\
\hline
{\textbf{Solution Expression:} 6000$\div$5$\times$3 \quad \textbf{Solution: }3600}\\ 
\hline

{\textbf{Problem 3:} Uncle Jack spends 5$\%$ of his bank account to invest for the trust funds of States and 15$\%$ of the account for the shares of Apple Inc. The money he has spent on financial management is 12500 dollars. How much money is in Uncle Jack's account?}\\
\hline
{\textbf{Solution Expression:} 12500$\div$(5$\%$ + 15$\%$) \quad \textbf{Solution: }62500}\\ 
{\textbf{Wrong Solution Expression:} 12500$\div$ 5 $\%$ $\times 15 \%$ }\\ 
\hline

\end{tabular}

\caption{A typical math word examples of the spurious correlation.}
\label{tab:case0}
\end{table}



In this paper, we propose a \textbf{s}elf-\textbf{c}onsistent \textbf{r}easoning framework (called SCR) to solve MWPs tasks. We obtain a sub-network by pruning the roberta2tree model denoted as the source network. Our SCR model adaptively finds spurious correlative samples through pruning. Specifically, SCR encourages the models' prediction consistency through mutual learning \cite{zhang2018deep, liang2020many}, which can emphasize samples with inconsistent prediction distributions between the source network and sub-network. 



We summarize our main contributions: (1) We propose a novel self-consistent reasoning framework for MWPs to expose spurious correlative samples and correct them adaptively. (2) We conduct extensive experiments on two benchmark datasets (i.e., Math23k and Ape210k). The results demonstrate that our model performs significantly better than the strong baselines. 

\section{Methodology}
\label{sec:metho}

A math word problems (MWPs) can be denoted by a projection $F: W \mapsto Y$, where ${W}=\{w_1,w_2,\dots,w_m\}$ is the problem sequence with $m$ words and $Y=\{y_1,y_2,\dots,y_n\}$ is the solution expression of the problem with $n$ words. MWPs aim to establish a model $F$ which generates a correct solution expression $Y$ and calculates the correct answer for the problem $W$.



\begin{figure}
     \centering
     \includegraphics[width=0.4\textwidth]{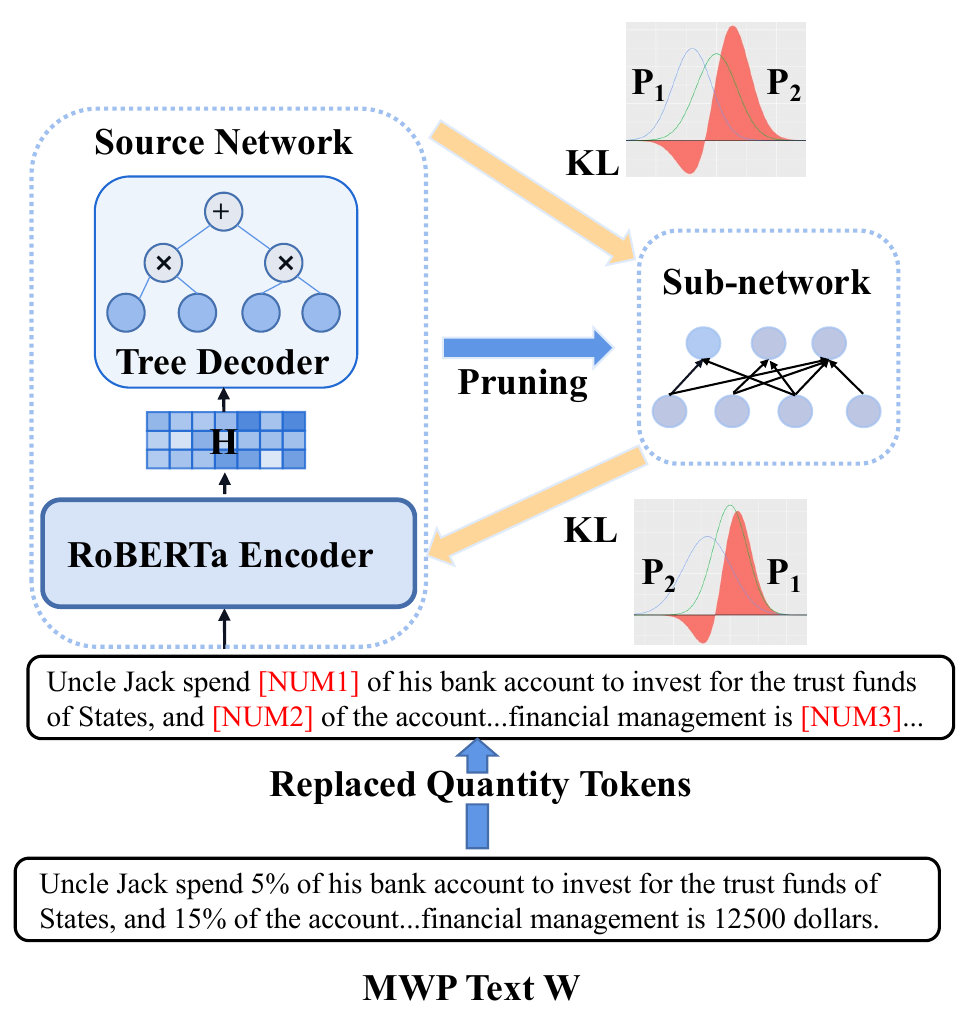}
     \caption{Overview of the proposed framework SCR} \label{fig2}
\end{figure}

As illustrated in Figure \ref{fig2}, the proposed SCR is composed of a source network (denoted as \emph{S}) and a sub-network (denoted as \emph{C}). The sub-network is obtained by pruning the source network. The two networks are optimized collaboratively and teach each other throughout the training process. We use the encoder-decoder framework as the backbone of both source and sub-networks. 


\subsection{The Encoder-Decoder Architecture}

To efficiently obtain a high-quality representation of the problem, we utilize the RoBERTa model \cite{liu2019roberta} as our encoder. We pass the problem sequence $W$ into the RoBERTa model and obtain problem representation $H \in \mathbb{R}^{m*d}$, where $d$ is the embedding size of the encoder. In order to model the relationship between the quantities in the pre-training model, we set up a learnable quantity embedding matrix $\emph{T}_E=\{t_1,t_2,\dots,t_n\}$, similar to the learnable position embedding in BERT \cite{devlin2018bert}. Before passing the sequence $W$ into the encoder, we first replace each quantity in the sequence $W$ with a token $t_i \in \emph{T}_t$. 

Inspired by GTS model \cite{xie2019goal}, our decoder uses the recursive operation of the decoder to construct $Y$ by order of pre-order traversal. First, the decoder generate the root node $t_{\rm root}$ (middle operator part) . Then, the decoder generates the left child node $t_{l}$. Furthermore, the right child node $t_{r}$ is generated. This process has been iterated until the leaf nodes are generated. Specifically, we apply the attention mechanism to learn the global context vector $G_i$, which is utilized to generate the current node token $\hat{y}_i$. Here we denote the digital embedding after being encoded by the encoder as $T$. The formula of the attention mechanism is shown below:
    \begin{equation}
    G_i=
    \left\{
    \begin{array}{lll}
    \text { Attention }( H, \emph{t}_{\rm root}, \emph{t}_l), \ \ \emph{t}_l  \notin \emptyset. \\
    \text { Attention }( H, \emph{t}_{\rm root}, \emph{t}_{sl}), \ \ \emph{t}_{sl} \notin \emptyset. \\
    \text { Attention }( H, \emph{t}_{\rm root}),\ \ \emph{t}_l, 
    \emph{q}_{sl} \in \emptyset.
    \end{array}
    \right.
    \end{equation}
\begin{equation}
    \hat{y}_i = {\rm Predict}(G_i, T).
\end{equation} 
where ${\rm Predict}(\cdot)$ is the final prediction layer for producing the tree node. 
    
The tree decoder will generate the left and right child nodes and push them into the stack using the top-down approach if the current node is an operator. Moreover, if it is a number, the merge operation will be carried out until the stack's leaf nodes emerge, at which point the result will be pushed into the stack of left child nodes for the attention operation. Then, the merge operation will pop the required node $\emph{t}_{\rm op}$ and $\emph{t}_{\rm subtree}$ from an embedding stack. The formula of recursive construction is as follows:    
\begin{gather}
    t_l = {\rm Left}(G_i, \hat{y_i}, t_{\rm root}).\\
    t_r = {\rm Right}(G_i, \hat{y_i}, t_{\rm root}).\\
    t_m = {\rm Merge}(t_{\rm op}, t_{\rm subtree}, t_{m-1}).
\end{gather}

\subsection{Self-consistent Reasoning} 

As shown in Figure \ref{fig2}, the proposed SCR comprises a source network and a sub-network. The sub-network is obtained by pruning the source network. In each iteration, the source network will correct the distribution shift of the output $p_{2}$ from the sub-network to implicitly emphasizes the spurious correlative samples. When we finish training the sub-network in this iteration, the sub-network also provides the supervision signals to correct the distribution shift from the output $p_{1}$ from the source network. Specifically, we preferentially fix the output distribution of the sub-network when neither network is trained by samples to expose spurious collaborative samples better. At the same time, these two networks are also trained by ground-truth supervision signals.


 
Formally, the training objective of the source network is to minimize negative log-likelihood (NLL) loss for each instance $(W,Y)$ from training data:
\begin{equation}
\small
\mathcal{L}_{S}(\theta_S)=-\sum_{i=1}^{n} \boldsymbol{y}_{i} \log p(\hat{\boldsymbol{y}_{i}} \mid W ; \theta_S).
\end{equation}
where $y_i$ is ground-truth of step i. $\theta_S$ denotes the parameters of the source network.  

We prune the model parameters $\theta_S$ of the source model and obtain the parameters $\theta_C$ of the sub-network. The training objective of the sub-network $C$ can be defined as:
\begin{equation}
\small
\mathcal{L}_{\rm C}(\theta_C)=-\sum_{i=1}^{n} \boldsymbol{y}_{i} \log p(\hat{\boldsymbol{y}_{i}} \mid W ; \theta_C).
\end{equation}

Inspired by the mutual learning~\cite{zhang2018deep, liang2020large}, we train the sub-network and the source network collaboratively by the symmetric Kullback Leibler (KL) Divergence.
First, we use the KL Divergence to measure the distance from the source network's prediction $\boldsymbol{p}_{1}$ to the sub-networks prediction $\boldsymbol{p}_{2}$ by:

\begin{equation}
\small
D_{K L}\left(\boldsymbol{p}_{2} \| \boldsymbol{p}_{1}\right)=\sum_{i=1}^{n} p_{2}\left(\hat{\boldsymbol{y}_{i}}\right) \log \frac{p_{2}\left(\hat{\boldsymbol{y}_{i}}\right)}{p_{1}\left(\boldsymbol{y}_{i}\right)}.
\label{equ:KL1}
\end{equation}
where $n$ is the length of the solution expression. $\boldsymbol{y}_{i}$ and $\hat{\boldsymbol{y}}_{i}$ denote the $i$-th ground-truth and generated tokens, respectively. Considering that KL divergence is asymmetric, we also calculate the divergence from $\emph{p}_{2}$ to $\emph{p}_{1}$:
\begin{equation}
\small
D_{K L}\left(\boldsymbol{p}_{1} \| \boldsymbol{p}_{2}\right)=\sum_{i=1}^{n} p_{1}\left(\hat{\boldsymbol{y}_{i}}\right) \log \frac{p_{1}\left(\hat{\boldsymbol{y}_{i}}\right)}{p_{2}\left(\boldsymbol{y}_{i}\right)} .
\label{equ:KL2}
\end{equation}

By averaging equations \ref{equ:KL1} and \ref{equ:KL2}, we get a symmetric KL divergence, denoted as $\bar{D}_{KL}$. Note that we alternately optimize $S$ and $C$ in each iteration, and calculate symmetric KL divergence $\bar{D}_{KL_1}$ and $\bar{D}_{KL_2}$  for $S$ and $C$, respectively.
We define the overall loss functions $\mathcal{L}_{S}$ and $\mathcal{L}_{C}$ for networks $S$ and $C$ respectively as follows:
\begin{gather}
\mathcal{L}_{S}=\mathcal{L}_{1}(\theta_S)+ \alpha\times \bar{D}_{KL_1}.\\
\mathcal{L}_{C}=\mathcal{L}_{2}(\theta_C)+\alpha\times \bar{D}_{KL_2}.
\end{gather}
where $\alpha$ is a proportional coefficient. 
In this way each network learns  to both correctly predict the ground-truth of training instances and  match the probability estimation of its peer network.

\section{Experiment}
\subsection{Experimental Setup}
\subsubsection{Datasets}
We conduct experiments on two benchmark MWPs datasets: Math23k \cite{wang2017deep} and Ape210k \cite{zhao2020ape210k}. 
Math23k contains 22162/1000 questions for training/testing, respectively. 
Ape210k is composed of 166,270 questions for training, 4,157 questions for validation, and 4,159 questions for testing.

\subsubsection{Implementation Details}
The word embedding size of the decoder is set to 1024. We adopt RoBERTa \cite{liu2019roberta} as the problem encoder. Following Roberta's setting, the encoder's hidden size is 768, and we set the hidden size of the decoder to 1024. We used Adamw \cite{loshchilov2018fixing} as the optimizer with the learning rate as 5e-5. The mini-batch size is set to 16. We adopt a beam search with a size of 5. Dropout (dropout rate = 0.5) is employed to avoid overfitting. For Ape210K, we set the maximum sequence length of questions as 150 and that of solution expressions as 50, similar to \cite{wu2021math}. For convergence, our model takes 80 epochs on Math23k and 50 epochs on Ape210k.


\subsubsection{Baselines}

\textbf{NS-Solver} \cite{qin2021neural}. This model uses auxiliary tasks to explicitly and seamlessly merge different levels of symbolic constraints to enhance the model's ability to solve MWPs.
\textbf{NumS2T} \cite{wu2021math}. This model uses the numerical properties prediction mechanism to capture the category and comparison information of the numerals and measure their importance in global expressions, which explicitly merges the numerical values into the sequence-to-tree network.
\textbf{TSN-MD} \cite{zhang2020teacher}. This model proposes a multi-decoder network based on distillation learning to generate diverse expressions.
\textbf{MATH-EN} \cite{wang2018translating}. This model proposes an equation normalization method to normalize repeated equations and also designs an ensemble method to improve the model's performance.
\textbf{Graph2Tree} \cite{zhang2020graph}. This model combines the advantages of graph-based encoders and tree-based decoders to generate better solution expressions.
\textbf{Multi-E/D} \cite{shen2020solving}. This model combines sequence-based encoders and graph-based encoders to enhance the presentation capabilities of text descriptions and generates different equation expressions through sequence-based decoders and tree-based decoders.
\textbf{GTS} \cite{xie2019goal}. This model designs a tree decoder to generate expressions.
\textbf{Tree-Decoder} \cite{liu2019tree}. This model proposes a tree-structured decoding method to generate abstract syntax trees of equations in a top-down manner.
\textbf{Ape} \cite{zhao2020ape210k}. This model designs a seq2seq model to solve the MWPs problem.
\textbf{StackDecoder} \cite{chiang2018semantically}. This model designs a stack-based decoder to generate expressions.
\textbf{KAS2T} \cite{wu2021math}. This model inserts external knowledge and state aggregation mechanisms into the seq2tree model.
\textbf{GenerateRank}~\cite{Shen2021GenerateR}.  This model devises a new ranking task for MWPs and proposes joint training with generation and ranking on a generative pre-trained language model.
\textbf{DeductReasoner}~\cite{Jie2022LearningTR}. This model views the task as a complex relation extraction problem and propose a novel approach that presents explainable deductive reasoning steps to iteratively construct target expressions.

\subsection{Experimental Results}

\subsubsection{Main Results}
The evaluation metric is answer accuracy. Table \ref{tab:accuracyS} show the performance comparison of our model with baseline methods on Math23K and Ape210k, respectively. Both our source network and sub-network achieve substantially better performance than the strong competitors, verifying the effectiveness of our self-consistent reasoning framework. We compute the accuracy of the generated solution expression. We consider an expression as a correct one when the predicted expression exactly matches the annotated solution. The expression prediction accuracy is reported in Table \ref{tab:accuracyE}.  We can see that the accuracy of solution expression generation is lower than the final answer prediction accuracy, showing that our model can generate some diverse solution expressions (not included in ground-truth expressions), leading to correct answers.
This experimental result also shows the strong generalization ability of our model because only after the model understands the original text can it predict equivalent and correct expressions instead of relying on some shallow heuristics in the original text.





\begin{table}
\centering

\resizebox{0.5\textwidth}{!}{
\begin{tabular}{cccc}
\hline
\textbf{Models} & \textbf{Math23k} & $\textbf{Math23k}^{\dagger}$ & \textbf{Ape210k}\\
\hline
{StackDecoder} & {-} & {65.8}  & {52.2}\\
{Tree-Decoder} & {69.0} & {-}  & {66.5}\\
{GTS} & {75.6} & {74.3}  & {67.7}\\ 
{KAS2T} & {76.3} & {-}  & {68.7}\\
{TSN-MD} & {77.4} & {75.1} & {-}\\ 
{Graph2Tree} & {77.4} & {75.5} & {-}\\ 
{NS-Solver} & {-} & {75.6} & {-}\\
{Ape} & {-} & {77.5}  & {70.2}\\
{NumS2T} & {78.1} & {-} & {70.5}\\
{Multi-E/D} & {78.4} & {76.9} & {-}\\

{GenerateRank} & {85.4} & {84.3} & {-}\\
{DeductReasoner} & {85.1} & {83.0} & {-}\\

\hline
{Source Network} & \textbf{85.5} & \textbf{84.5}  & \textbf{76.3}\\
{Sub-network} & \textbf{86.8} & \textbf{84.6}  & \textbf{76.7}\\
\hline
\end{tabular}
}
\caption{Solution accuracy of SCR and various baselines. Note
that Math23K denote results on public test set and $\text{Math23K}^{\dagger}$ denote
5-fold cross-validation. }
\label{tab:accuracyS}
\end{table}

\begin{table}
\centering
\resizebox{0.4\textwidth}{!}{
\begin{tabular}{ccc}
\hline
\textbf{Models} & \textbf{Answer-ac} & \textbf{Equation-ac} \\
\hline

{MATH-EN} & {66.7} & {60.1}\\
{GTS} & {75.6} & {64.8}\\ 
{TSN-MD} & {77.4} & {65.8}\\

\hline
{Source Network} & \textbf{85.5} & \textbf{73.3}\\
{Sub-network} & \textbf{86.8} & \textbf{73.5}\\
\hline
\end{tabular}
}
\caption{Accuracy of equation generation on Math23k. }
\label{tab:accuracyE}
\end{table}

\begin{table}
\centering
\resizebox{0.4\textwidth}{!}{
\begin{tabular}{cc}
\hline
\textbf{Models} & \textbf{Math23k} \\
\hline
{Source network w/o  pruning} & {84.8}\\
{Sub-network w/o pruning} & {85.6}\\
\hline
{Source network w/o MT} & {84.2}\\
{Sub-network w/o MT } & {84.3}\\
\hline
{Source Network} & \textbf{85.5} \\
{Sub-network} & \textbf{86.8} \\
\hline
\end{tabular}
}
\caption{Ablation study on Math23k. }
\label{tab:ablation}
\end{table}

\begin{table}
\centering
\resizebox{0.4\textwidth}{!}{
\begin{tabular}{cc|cc}
\hline
$\boldsymbol{\alpha}$ & \textbf{Math23k} & \textbf{Pruning Rate}& \textbf{Math23k}\\
\hline
{0.0005} & {85.1} & {0.1} & {85.8}\\
{0.005} & {86.8} & {0.2} & {86.8}\\ 
{0.05} & {$\triangle$} & {0.3} & {84.4}\\ 
{-} & {-} & {0.4} & {84.4}\\
{-} & {-} & {0.5} & {$\triangle$}\\
\hline
\end{tabular}}
\caption{The Sensitivity Analysis of $\alpha$ and Pruning Rate. $\triangle$
denote divergence.}
\label{tab:alpha}
\end{table}

\begin{table}
\small
\centering
\begin{tabular}{l|p{5.5cm}}

\hline
{Problem Text A:} & {John has a fixed amount of money and saves 50 dollars every day in his bank account by 25 days. If it takes 20 days to complete the process, how many money will he save per day? }\\
\hline
{Ground-truth:} & {$50 \times 25 \div 20$}\\ 
{Source Network:} & {$50 \times 25 \div 20$}\\ 
{Sub-network:} & {$25 \div 20 \times 50$}\\ 
\hline

{Problem Text B:} & {Alice gets the money from the bank's account by 5 days. After that, Alice takes another 3 days to get the money from the same account again. If she takes 65 dollars per day. How much is she takes out from the bank? }\\
\hline
{Ground-truth:} & {(5 $+$ 3) $\times$ 65}\\ 
{Source Network:} & {(5 $+$ 3) $\times$ 65}\\ 
{Sub-network:} & {5 $\times$ 65 $+$ 3 $\times$ 65}\\ 
\hline

\end{tabular}
\caption{Case study from Math23k to solve spurious correlation.}
\label{tab:case}
\end{table}

\subsubsection{Ablation Study}
We conduct an ablation test on Math23k to analyze the impact of different components in SCR. First, we remove the mutual learning from the source network and sub-network, denoted as source network w/o MT and sub-network w/o MT, respectively. Second, we replace the pruned sub-network with the source network to evaluate the impact of pruning (denoted as source network w/o pruning and sub-network w/o pruning, respectively). We summarize the results in Table. \ref{tab:ablation}.
Both the pruning strategy and mutual learning contribute greatly to the performance of SCR.  

\subsubsection{The Sensitivity Analysis of $\boldsymbol{\alpha}$} 
We analyze the sensitivity of $\alpha$  on Math23k. As shown in Table \ref{tab:alpha}, when $\alpha$ is greater than or equal to 0.05, the sub-network will not converge. We obtain the best result when $\alpha=0.005$.

\subsubsection{The Sensitivity Analysis of Pruning Rate} 
We analyze the sensitivity of pruning rate on the competitor network. As shown in Table \ref{tab:alpha}, when the pruning rate is greater than 0.2 (the best value), the performance of the sub-network drops quickly.

\subsubsection{Case Study}
As an intuitive way to show the performance of SCR, we randomly choose two problems with similar semantic information from the dataset and show its solution expression generated by our model. As shown in Table \ref{tab:case}, we observe that our model can produce two different but equivalent solutions for each problem. Specifically, it shows that our model has learned the solutions' equivalence, which implies our model really captures the logic of the original text rather than relying on shallow heuristics. Thus, it can efficiently alleviate spurious correlation of the problems containing similar semantic information (e.g., \textit{bank of account, money, per day}), which may easily cause the wrong answer like Table~\ref{tab:case0}.


\section{Conclusion}
\label{sec:conclusion}
In this paper, we proposed a self-consistent reasoning framework to solve MWPs tasks. Our SCR model implicitly corrects spurious correlative samples cooperatively learning a source network and a pruned sub-network. Extensive experiments on two benchmark MWPs datasets demonstrated the effectiveness of our model. 

In addition, the effect of the learnable quantity embedding we proposed is also significant, which can prevent the model from using the commutative law incorrectly.  Experiment results show that our proposal has achieved the most advanced results on the Math23k and Ape210k datasets. At the same time, our method also verified the correctness of the lottery hypothesis in the MWP task. In our future work, we will consider introducing negative samples to the model learning in order to distinguish between equivalent expressions and similar but not equivalent expressions.


\clearpage
\bibliographystyle{IEEEbib}
\bibliography{refs}

\begin{thebibliography}{10}

\bibitem{bobrow1964natural}
Daniel~G Bobrow,
\newblock ``Natural language input for a computer problem solving system,''
\newblock 1964.

\bibitem{wei2022chain}
Jason Wei, Xuezhi Wang, Dale Schuurmans, Maarten Bosma, Ed~Chi, Quoc Le, and Denny Zhou,
\newblock ``Chain of thought prompting elicits reasoning in large language models,''
\newblock {\em arXiv preprint arXiv:2201.11903}, 2022.

\bibitem{wang2022self}
Xuezhi Wang, Jason Wei, Dale Schuurmans, Quoc Le, Ed~Chi, and Denny Zhou,
\newblock ``Self-consistency improves chain of thought reasoning in language models,''
\newblock {\em arXiv preprint arXiv:2203.11171}, 2022.

\bibitem{li2022advance}
Yifei Li, Zeqi Lin, Shizhuo Zhang, Qiang Fu, Bei Chen, Jian-Guang Lou, and Weizhu Chen,
\newblock ``On the advance of making language models better reasoners,''
\newblock {\em arXiv preprint arXiv:2206.02336}, 2022.

\bibitem{Wan2022GMAPGM}
Zhongwei Wan, Yichun Yin, Wei Zhang, Jiaxin Shi, Lifeng Shang, Guangyong Chen, Xin Jiang, and Qun Liu,
\newblock ``G-map: General memory-augmented pre-trained language model for domain tasks,''
\newblock in {\em Conference on Empirical Methods in Natural Language Processing}, 2022.

\bibitem{wang2024iot}
Xin Wang, Zhongwei Wan, Arvin Hekmati, Mingyu Zong, Samiul Alam, Mi~Zhang, and Bhaskar Krishnamachari,
\newblock ``Iot in the era of generative ai: Vision and challenges,''
\newblock {\em arXiv preprint arXiv:2401.01923}, 2024.

\bibitem{wan2023efficient}
Zhongwei Wan, Xin Wang, et~al.,
\newblock ``Efficient large language models: A survey,''
\newblock {\em arXiv preprint arXiv:2312.03863}, 2023.

\bibitem{shen2021generate}
Jianhao Shen, Yichun Yin, Lin Li, Lifeng Shang, Xin Jiang, Ming Zhang, and Qun Liu,
\newblock ``Generate \& rank: A multi-task framework for math word problems,''
\newblock {\em arXiv preprint arXiv:2109.03034}, 2021.

\bibitem{wan2023text}
Zhongwei Wan,
\newblock ``Text classification: A perspective of deep learning methods,''
\newblock {\em arXiv preprint arXiv:2309.13761}, 2023.

\bibitem{huang2021recall}
Shifeng Huang, Jiawei Wang, Jiao Xu, Da~Cao, and Ming Yang,
\newblock ``Recall and learn: A memory-augmented solver for math word problems,''
\newblock {\em arXiv preprint arXiv:2109.13112}, 2021.

\bibitem{wang2018translating}
Lei Wang, Yan Wang, Deng Cai, Dongxiang Zhang, and Xiaojiang Liu,
\newblock ``Translating a math word problem to an expression tree,''
\newblock {\em arXiv preprint arXiv:1811.05632}, 2018.

\bibitem{wan2023spatio}
Zhongwei Wan, Xin Liu, Benyou Wang, Jiezhong Qiu, Boyu Li, Ting Guo, Guangyong Chen, and Yang Wang,
\newblock ``Spatio-temporal contrastive learning-enhanced gnns for session-based recommendation,''
\newblock {\em ACM Transactions on Information Systems}, vol. 42, no. 2, pp. 1--26, 2023.

\bibitem{liu2023etp}
Che Liu, Zhongwei Wan, Sibo Cheng, Mi~Zhang, and Rossella Arcucci,
\newblock ``Etp: Learning transferable ecg representations via ecg-text pre-training,''
\newblock {\em arXiv preprint arXiv:2309.07145}, 2023.

\bibitem{zhang2020graph}
Jipeng Zhang, Lei Wang, Roy Ka-Wei Lee, Yi~Bin, Yan Wang, Jie Shao, and Ee-Peng Lim,
\newblock ``Graph-to-tree learning for solving math word problems,''
\newblock Association for Computational Linguistics, 2020.

\bibitem{xiong2022expression}
Jing Xiong, Chengming Li, Min Yang, Xiping Hu, and Bin Hu,
\newblock ``Expression syntax information bottleneck for math word problems,''
\newblock in {\em Proceedings of the 45th International ACM SIGIR Conference on Research and Development in Information Retrieval}, 2022, pp. 2166--2171.

\bibitem{jie2022learning}
Zhanming Jie, Jierui Li, and Wei Lu,
\newblock ``Learning to reason deductively: Math word problem solving as complex relation extraction,''
\newblock {\em arXiv preprint arXiv:2203.10316}, 2022.

\bibitem{patel2021nlp}
Arkil Patel, Satwik Bhattamishra, and Navin Goyal,
\newblock ``Are nlp models really able to solve simple math word problems?,''
\newblock {\em arXiv preprint arXiv:2103.07191}, 2021.

\bibitem{kumar2021adversarial}
Vivek Kumar, Rishabh Maheshwary, and Vikram Pudi,
\newblock ``Adversarial examples for evaluating math word problem solvers,''
\newblock {\em arXiv preprint arXiv:2109.05925}, 2021.

\bibitem{hooker2019compressed}
Sara Hooker, Aaron Courville, Gregory Clark, Yann Dauphin, and Andrea Frome,
\newblock ``What do compressed deep neural networks forget?,''
\newblock {\em arXiv preprint arXiv:1911.05248}, 2019.

\bibitem{jiang2021self}
Ziyu Jiang, Tianlong Chen, Bobak~J Mortazavi, and Zhangyang Wang,
\newblock ``Self-damaging contrastive learning,''
\newblock in {\em International Conference on Machine Learning}. PMLR, 2021, pp. 4927--4939.

\bibitem{zhang2018deep}
Ying Zhang, Tao Xiang, Timothy~M Hospedales, and Huchuan Lu,
\newblock ``Deep mutual learning,''
\newblock in {\em Proceedings of the IEEE conference on computer vision and pattern recognition}, 2018, pp. 4320--4328.

\bibitem{liang2020many}
Zhenyu Liang, Yunfan Li, and Zhongwei Wan,
\newblock ``Many-objective estimation of distribution optimization algorithm based on wgan-gp,''
\newblock {\em arXiv preprint arXiv:2003.08295}, 2020.

\bibitem{liu2019roberta}
Yinhan Liu, Myle Ott, Naman Goyal, Jingfei Du, Mandar Joshi, Danqi Chen, Omer Levy, Mike Lewis, Luke Zettlemoyer, and Veselin Stoyanov,
\newblock ``Roberta: A robustly optimized bert pretraining approach,''
\newblock {\em arXiv preprint arXiv:1907.11692}, 2019.

\bibitem{devlin2018bert}
Jacob Devlin, Ming-Wei Chang, Kenton Lee, and Kristina Toutanova,
\newblock ``Bert: Pre-training of deep bidirectional transformers for language understanding,''
\newblock {\em arXiv preprint arXiv:1810.04805}, 2018.

\bibitem{xie2019goal}
Zhipeng Xie and Shichao Sun,
\newblock ``A goal-driven tree-structured neural model for math word problems,''
\newblock in {\em IJCAI}, 2019, pp. 5299--5305.

\bibitem{liang2020large}
Zhenyu Liang, Yunfan Li, and Zhongwei Wan,
\newblock ``Large scale many-objective optimization driven by distributional adversarial networks,''
\newblock {\em arXiv preprint arXiv:2003.07013}, 2020.

\bibitem{wang2017deep}
Yan Wang, Xiaojiang Liu, and Shuming Shi,
\newblock ``Deep neural solver for math word problems,''
\newblock in {\em Proceedings of the 2017 Conference on Empirical Methods in Natural Language Processing}, 2017, pp. 845--854.

\bibitem{zhao2020ape210k}
Wei Zhao, Mingyue Shang, Yang Liu, Liang Wang, and Jingming Liu,
\newblock ``Ape210k: A large-scale and template-rich dataset of math word problems,''
\newblock {\em arXiv preprint arXiv:2009.11506}, 2020.

\bibitem{loshchilov2018fixing}
Ilya Loshchilov and Frank Hutter,
\newblock ``Fixing weight decay regularization in adam,''
\newblock 2018.

\bibitem{wu2021math}
Qinzhuo Wu, Qi~Zhang, Zhongyu Wei, and Xuan-Jing Huang,
\newblock ``Math word problem solving with explicit numerical values,''
\newblock in {\em Proceedings of the 59th Annual Meeting of the Association for Computational Linguistics and the 11th International Joint Conference on Natural Language Processing (Volume 1: Long Papers)}, 2021, pp. 5859--5869.

\bibitem{qin2021neural}
Jinghui Qin, Xiaodan Liang, Yining Hong, Jianheng Tang, and Liang Lin,
\newblock ``Neural-symbolic solver for math word problems with auxiliary tasks,''
\newblock {\em arXiv preprint arXiv:2107.01431}, 2021.

\bibitem{zhang2020teacher}
Jipeng Zhang, Ka~Wei LEE, Ee-Peng Lim, Wei Qin, Lei Wang, Jie Shao, Qianru Sun, et~al.,
\newblock ``Teacher-student networks with multiple decoders for solving math word problem,''
\newblock 2020.

\bibitem{shen2020solving}
Yibin Shen and Cheqing Jin,
\newblock ``Solving math word problems with multi-encoders and multi-decoders,''
\newblock in {\em Proceedings of the 28th International Conference on Computational Linguistics}, 2020, pp. 2924--2934.

\bibitem{liu2019tree}
Qianying Liu, Wenyv Guan, Sujian Li, and Daisuke Kawahara,
\newblock ``Tree-structured decoding for solving math word problems,''
\newblock in {\em Proceedings of the 2019 Conference on Empirical Methods in Natural Language Processing and the 9th International Joint Conference on Natural Language Processing (EMNLP-IJCNLP)}, 2019, pp. 2370--2379.

\bibitem{chiang2018semantically}
Ting-Rui Chiang and Yun-Nung Chen,
\newblock ``Semantically-aligned equation generation for solving and reasoning math word problems,''
\newblock {\em arXiv preprint arXiv:1811.00720}, 2018.

\bibitem{Shen2021GenerateR}
Jianhao Shen, Yichun Yin, Lin Li, Lifeng Shang, Xin Jiang, Ming Zhang, and Qun Liu,
\newblock ``Generate \& rank: A multi-task framework for math word problems,''
\newblock {\em ArXiv}, vol. abs/2109.03034, 2021.

\bibitem{Jie2022LearningTR}
Zhanming Jie, Jierui Li, and Wei Lu,
\newblock ``Learning to reason deductively: Math word problem solving as complex relation extraction,''
\newblock in {\em ACL}, 2022.

\end{thebibliography}

\end{document}